\documentclass{article}
\usepackage{ijcai24}

\usepackage{times}
\usepackage{soul}
\usepackage{url}
\usepackage[hidelinks]{hyperref}
\usepackage[utf8]{inputenc}
\usepackage[small]{caption}
\usepackage{graphicx}
\usepackage{amsmath}
\usepackage{amsthm}
\usepackage{booktabs}
\usepackage{algorithm}
\usepackage{algorithmic}
\usepackage[switch]{lineno}

\usepackage{wrapfig}
 \usepackage{amssymb}
 \usepackage{dirtytalk}
 \usepackage{multirow}
\usepackage{subcaption}

\usepackage{xcolor}         

\usepackage{newfloat}
\usepackage{listings}
\usepackage{amsfonts,amsmath}
\usepackage{bm}

\usepackage{tikz}
\usetikzlibrary{positioning,decorations.pathreplacing,shapes}

\newcommand{\digit}[1]{\vcenter{\hbox{\includegraphics[height=10pt]{#1}}}}

\newcommand{\Sym}{\mathcal{S} }

\title{Simple and Effective Transfer Learning for Neuro-Symbolic Integration}

\author{
Alessandro Daniele$^{1*}$
\and
Tommaso Campari$^{1}$\footnote{Equal Contribution}
\and
Sagar Malhotra$^{2}$\And
Luciano Serafini$^1$\\
\affiliations
$^1$Fondazione Bruno  Kessler, Italy\\
$^2$TU Wien, Austria\\
\emails
}

\begin{document}

\maketitle

\begin{abstract}
Deep Learning (DL) techniques have achieved remarkable successes in recent years. However, their ability to generalize and execute reasoning tasks remains a challenge. A potential solution to this issue is Neuro-Symbolic Integration (NeSy), where neural approaches are combined with symbolic reasoning. Most of these methods exploit a neural network to map perceptions to symbols and a logical reasoner to predict the output of the downstream task. These methods exhibit superior generalization capacity compared to fully neural architectures. However, they suffer from several issues, including slow convergence, learning difficulties with complex perception tasks, and convergence to local minima. This paper proposes a simple yet effective method to ameliorate these problems.  The key idea involves pretraining a neural model on the downstream task. Then, a NeSy model is trained on the same task via transfer learning, where the weights of the perceptual part are injected from the pretrained network. The key observation of our work is that the neural network fails to generalize only at the level of the symbolic part while being perfectly capable of learning the mapping from perceptions to symbols. We have tested our training strategy on various SOTA NeSy methods and datasets, demonstrating consistent improvements in the aforementioned problems.
\end{abstract}

    \section{Introduction}
    \begin{figure*}[t]
   \centering
   \includegraphics[width=\linewidth]{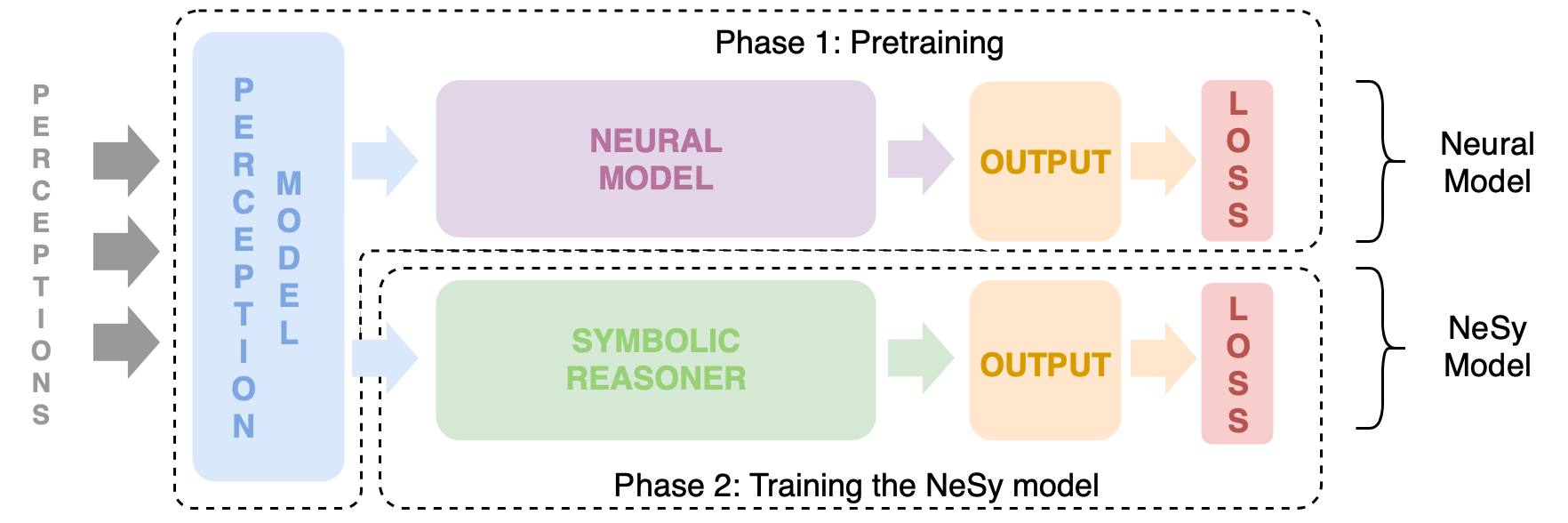}
   \caption{Training procedure overview: in the first phase, a neural network model is trained on the downstream task; in the second phase, the NeSy method is trained starting from the previously learned perception model.}
   \label{fig:teaser}
\end{figure*}

Methods based on Neural Networks (NNs) have advanced the state-of-the-art across a wide array of fields like image recognition~\cite{image_recognition}, speech processing~\cite{hinton2012deep_speech}, and natural language processing~\cite{DL_NLP}. However, even in applications where Deep Learning (DL) excels, it shows minimal reasoning capabilities~\cite{lievin2022can_LLm}. This severely restricts the applicability of NNs in scenarios where reasoning over perception inputs is required to guarantee safety and reliability, e.g., in  autonomous driving applications~\cite{giunchiglia2023road_R}. \emph{Neuro-Symbolic} (NeSy)~\cite{NeSy_Survey,Nesy_2021}  systems  integrate  NNs and symbolic reasoning. Using a reasoning engine, they allow learning of symbolic abstraction for perception inputs and reasoning over these symbolic abstractions. A key feature of NeSy systems is that they support explainability and out-of-distribution generalization for complex tasks. For example, in the MNIST addition task~\cite{manhaeve2018deepproblog}, as shown in Figure~\ref{fig:method} (right), a NeSy system can learn the label $3$ and $0$ for the MNIST images, given only the fact that the MNIST digits sum to $3$, and the rule that $3+ 0=3$. Since an expert can change the rules in the symbolic reasoner, the same architecture can be utilized for multi-digit addition tasks, where two sequences of MNIST digits must be added. This can be achieved by only providing new expert rules with no additional training required for the perception part. Furthermore, recent lines of NeSy works, such as Deep Symbolic Learning (DSL)~\cite{DSL} and Deep Concept Reasoner (DCR)~\cite{Mara_Concept_Bottleneck}, have also managed to learn both the logical rules and the perception labels simultaneously --- showing that NeSy systems have the potential for out-of-distribution generalization even without any expert intervention.

Most NeSy systems comprise a perception part, mapping perception inputs to symbols, and a reasoning engine that gives the final output. The perception part is composed of one or more NNs. A key challenge to NeSy systems is the need to train such NNs through the weak supervision coming from a symbolic reasoning model. This is significantly challenging as the training signal needs to be propagated from discrete symbolic reasoners to continuously parameterized NNs. Consequently, the NeSy model could be affected by slow convergence and face difficulties in dealing with complex perception inputs. Furthermore, the weak supervision is provided in the form of labels of the downstream task. As an example,
the label $x+y=3$ of the downstream task (see Figure~\ref{fig:method}) can be satisfied by multiple labellings, e.g. $1+2 =3$, $0+3=3$ etc.  This issue has been formally characterized as \emph{Reasoning Shortcut} (RS)~\cite{marconato2023neuro} and can lead to stagnation in local minima. More precisely, an RS occurs when some latent concepts learned by a NeSy architecture do not correspond to the intended concept, i.e., the concept expected by the given knowledge. Our strategy could reduce this problem since it reduces the effort of the NeSy model for learning the concepts. One specific instance of this problem has been recognized in~\cite{ILR}, where Iterative Local Refinement (ILR) has been trapped in this kind of local minima for the MNISTSum task. For this reason, we tested the pretraining strategy with ILR, obtaining a consistent reduction of local minima.


Previous works show a significant improvement by adopting pre-processing techniques that learn perception labels separately from the reasoning task.  \cite{SAT_Net_Solve} uses generative adversarial neural networks (GANs)~\cite{goodfellow2014generative} to obtain MNIST digit clusters. Embed2sym~\cite{EMB2SYM} takes a different direction, it trains a fully neural model directly on the downstream task. It then extracts the embeddings for the perception inputs and uses K-Means clustering to cluster the embeddings. The problem of mapping the cluster labels to the intended logical symbols
is then solved using answer set programming. A key challenge with both of these methods is that the clustering process requires intervention from an expert for the choice of the clustering algorithm and the distance metric. 
Furthermore, it is hard to see how these methods can be integrated into or help end-to-end systems where  both rules and symbols are learnt simultaneously, e.g.,\cite{DSL,OOD_Generalization}. 
Instead, our framework is easily applicable in these types of methods, like DSL, where the symbolic rules are learned alongside the perceptions.

This paper proposes an adaptive pre-processing step that learns embeddings of the perception inputs based on the downstream task. Unlike the methods proposed in the literature, our framework does not require additional supervision and applies to a vast array of NeSy systems. It constitutes a simple pretraining step that learns informative embeddings for perception inputs, just using the supervision of the downstream task, without any additional clustering or symbolic reasoning. Figure~\ref{fig:teaser} shows a general overview of the method, where two models for the downstream tasks are defined. The neural model continuously approximates the symbolic reasoner and is trained in the first phase. The perception model, which maps input features to embeddings, is then used by the NeSy model in the second learning phase. Then, at inference, only the NeSy architecture is used since the symbolic reasoner perfectly represents the symbolic knowledge rather than approximating it, allowing for better out-of-distribution generalization than the neural architecture.

We empirically demonstrate that these embeddings form a much more tailored input for training the NeSy system than the raw input itself.  Our method significantly improves classical NeSy benchmarks and can also scale NeSy systems to previously unattainable, complex tasks.

    
    
    \section{Related Works}


Neuro-Symbolic Integration~\cite{NeSy_Survey,Nesy_2021,Raedt2020FromSR} have evolved in multiple directions. Some of the key differentiating features amongst NeSy systems are the type of logic they use, and the way in which they exploit knowledge.

A large variety of NeSy systems are based on the idea of enforcing logical constraints on the outputs of a NN. In this category, most methods include logical constraints through the addition of a regularization term in the loss, which enforces satisfaction of the logical constraints. Semantic loss~\cite{SL}  aims at guiding the NN training through a logic-based differentiable regularizer based on probabilistic semantics, obtained by compiling logical knowledge into a Sentential Decision Diagram (SDD) \cite{SDD}. Other methods in this category are Logic Tensor Networks~\cite{LTN} and Semantic-Based Regularization~\cite{SBR}, which encode logical knowledge into a differentiable function based on fuzzy logic semantics. This function term is then used as a regularizer in the loss function. Since these methods incorporate symbolic knowledge into the loss function, the knowledge encoded in the logical constraints does not play any role at inference time. Other methods that enforce constraints in NN outputs are Knowledge Enhanced Neural Network (KENN)~\cite{Daniele2019KnowledgeEN}, and Iterative Local Refinement (ILR)~\cite{ILR}. 
Unlike previously mentioned methods,
ILR and KENN extend a basic NN with additional layers that change the predictions to enhance logical knowledge satisfaction. As in LTN and SBR, the satisfaction level of a formula is computed using  fuzzy semantics. KENN and ILR exploit background knowledge also at inference time. Note that, unlike the systems based on probabilistic logic, fuzzy logic-based methods have the advantage that the exploration of the symbolic search space can take place without any additional cost of the potentially intractable knowledge compilation and weighted model counting steps. However, these systems tend to stagnate in local minima.



Another strand consists of NeSy methods that are obtained by extending existing symbolic reasoners with neural inputs. DeepProbLog~\cite{manhaeve2018deepproblog} is a neural extension of ProbLog~\cite{ProbLog}. 
The key feature of DeepProbLog is that it admits neural predicates i.e., predicates whose associated probability parameters are obtained as an output from a NN. 
The system utilizes SDD~\cite{SDD} fortified with gradient semirings, to provide an end-to-end differentiable system capable of learning the neural network through the weak supervision coming from the downstream task's labels. Recent advancements have offered similar neural extensions to other symbolic solvers. DeepStochLog~\cite{winters2022deepstochlog} extends Stochastic Definite Clause Grammars~\cite{stoch_grammer} --- a context-sensitive probabilistic grammar formalism. Similarly, NeurASP~\cite{yang2020neurasp} extends Answer Set Programming~\cite{ASP}. A key challenge faced by systems extending symbolic reasoners is that they inherit, and potentially worsen, the computational complexity of inference.


More recently, new NeSy approaches have been proposed that enable learning both the symbolic labels for perception inputs and the symbolic rules themselves. The first method introduced in this category is Deep Symbolic Learning (DSL)~\cite{DSL}, where Reinforcement Learning policies are used to discretize internal latent symbols and select the underlying symbolic rules. Although it learns also the rules, DSL proved to be more scalable than other NeSy frameworks that exploit the given knowledge. However, training DSL can be challenging due to the slow convergence when the number of internal symbols increases. For instance, to learn the multi-digit version of the sum, it is required to use curriculum learning. Our experimental analysis shows that our pretraining step allows us to improve the efficacy of DSL reducing the aforementioned problems, allowing for learning the multi-digit sum without curriculum learning and generally increasing the convergence speed.

A more recent method that learns both rules and perceptions is Deep Concept Reasoner (DCR)~\cite{Mara_Concept_Bottleneck}, which builds syntactic rule structures using concept embeddings. The concept embeddings have vectors of fuzzy truth values associated with them, leading to clear and interpretable semantics for learned rules, and subsequent inference performed on them. However, DCR requires more computational resources as compared to DSL and it has not been considered in our work.






The most relevant NeSy approach to our work is Embed2Sym~\cite{EMB2SYM}. It separately trains a NN on the downstream NeSy task and utilizes clustering for extracting symbolic concepts from the learned embeddings. It then uses Answer Set Programming(ASP) to map cluster labels to logical symbols based on the given symbolic knowledge. While similar to our pretraining strategy, the method is specialized on a specific architecture, and can not be applied on other methods. Conversely, our approach can be applied on a variety of NeSy systems, including the ones that learns the logical rules, allowing for solving tasks precluded to Embed2Sym.

    \section{Method}
\label{sec:method}
\begin{figure}
\centering
  \includegraphics[width=0.48\textwidth]{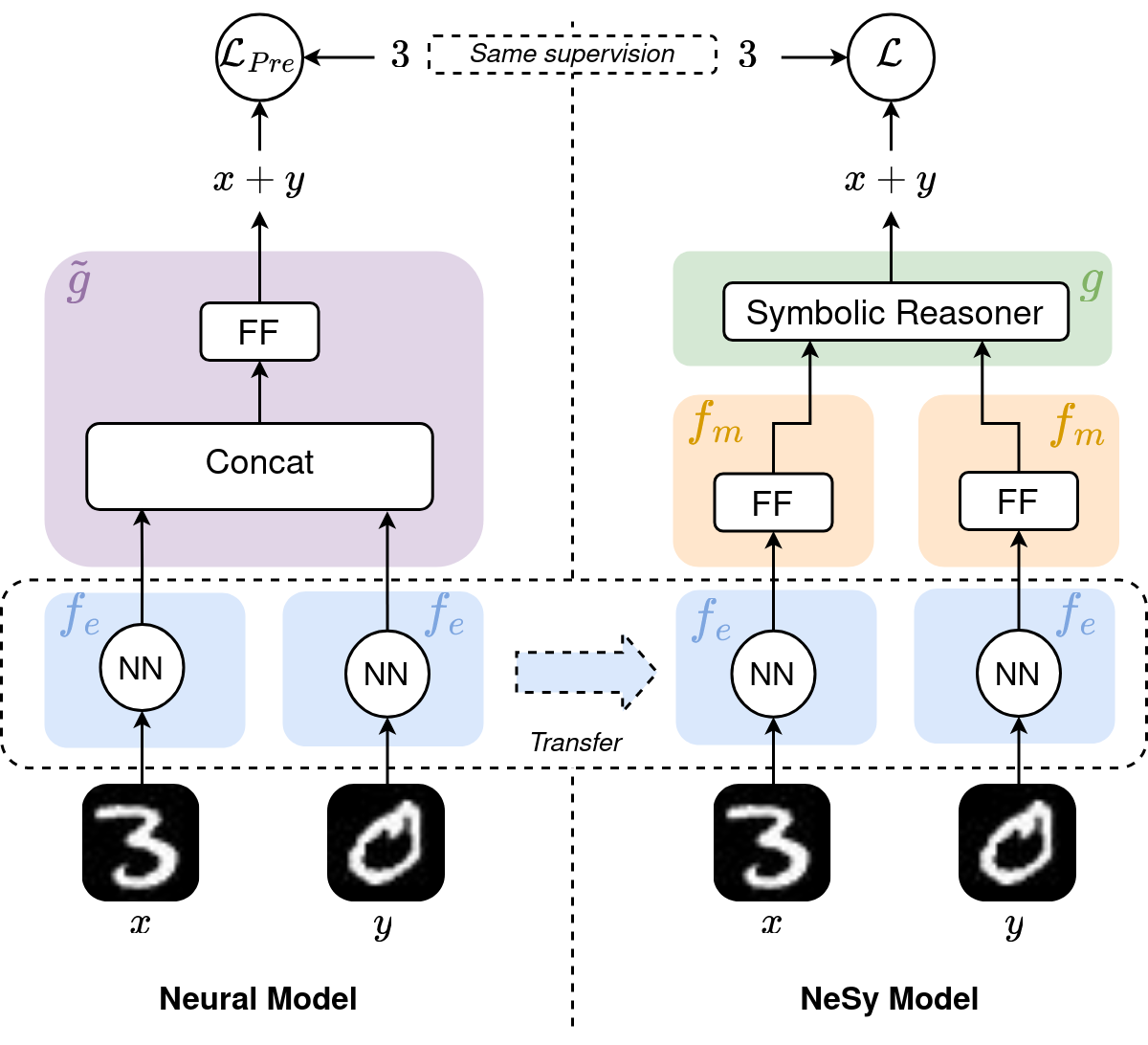}
    \caption{An example of our learning strategy on the MNISTSum task: on the first phase (left), we train a neural model for the downstream task; on the second phase (right), we use the pretrained weights of the neural network $f_e$ as a starting point for the NeSy architecture, which learns the mapping from embeddings to symbols ($f_m$) and, in case of DSL, the symbolic function $g$.}
    \label{fig:method}
\end{figure}

NeSy systems\footnote{
Here we refer to NeSy systems in the specific context where the symbolic reasoner is employed to infer new facts from the symbolic knowledge. This excludes methods like LTN where the knowledge is merely used to constrain the outputs of the neural network. It should be noted that not all NeSy systems operate in this manner.} can be seen as a composition of \emph{perception} and \emph{symbolic} functions~\cite{DSL}. 
In most NeSy systems, the perception function $f: \mathcal{X} \rightarrow \Sym$ is represented as a NN with learnable parameters, whereas $g:  \Sym^n \rightarrow \Sym$ represents expert-provided rules which allow to infer the downstream task labels from the internal latent symbols found by $f$~\cite{manhaeve2018deepproblog,winters2022deepstochlog,ILR}.
In other words, the NeSy system computes a downstream task that consists in the 
function   $g(f(x_1),\dots,f(x_n))$.

\begin{figure*}[t!]
\centering
    \begin{subfigure}{0.49\linewidth}
    \centering
        \includegraphics[width=\linewidth]{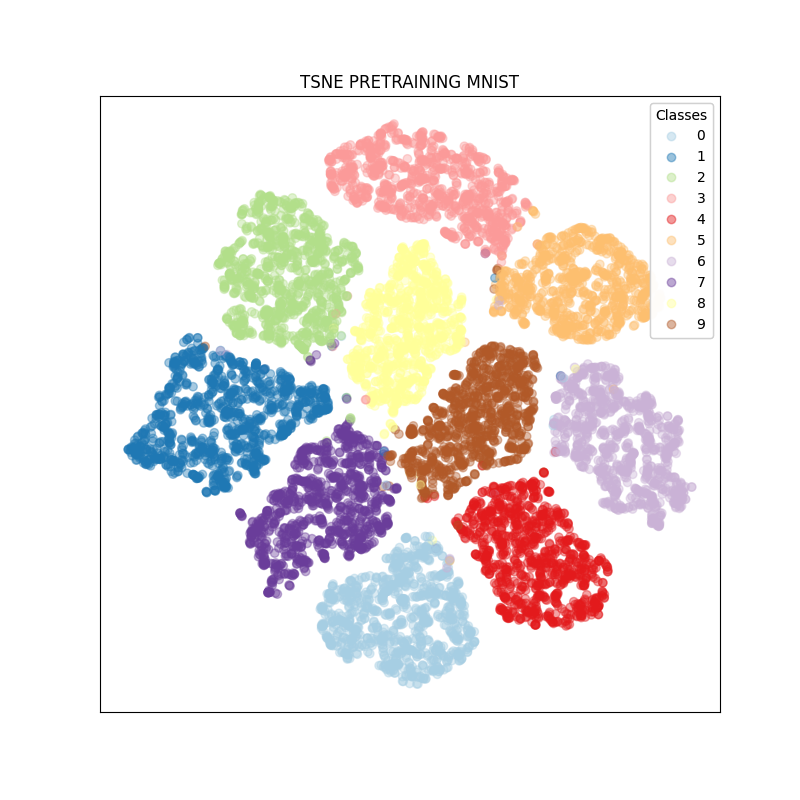}
        \vspace{-27pt}
        \caption{}
    \end{subfigure}
    \begin{subfigure}{0.49\linewidth}
    \centering
        \includegraphics[width=\linewidth]{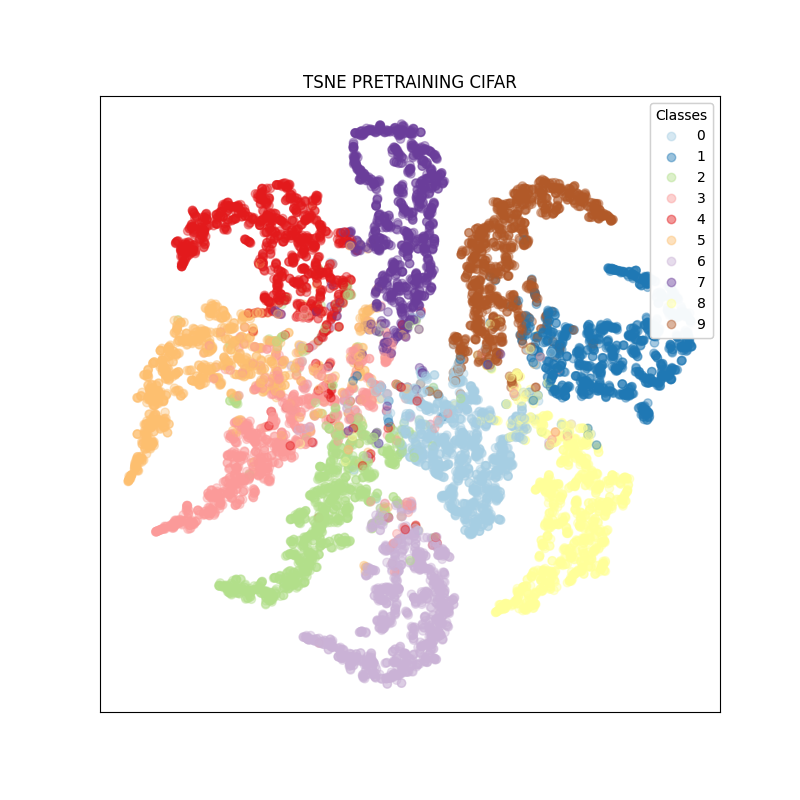}
        \vspace{-27pt}
        \caption{}
    \end{subfigure}
    \caption{t-SNE applied to embeddings of $f_e$ learned by the neural model on the MNISTSum (left) and CIFARSum (right) tasks. Colours represent different digits.}
    \label{fig:tsne}
\end{figure*}

Figure~\ref{fig:method} (right) shows the typical architecture of a NeSy system for the MNISTSum task, where two MNIST images $x$ and $y$ are provided as input, and the downstream task consists on calculating the sum of their respective digits. We additionally assume that the perception neural network $f$ is represented as the composition of two functions, $f_e$ (blue box) and $f_m$ (orange box), representing the mapping from perceptions to embeddings and the mapping from embeddings to symbols, respectively. Formally: 
$$f(x)=f_m(f_e(x))$$
where $f_e: \mathcal{X}^{n} \rightarrow \mathcal{R}^m$ and $f_m: \mathcal{R}^m \to \Sym$. The idea is to train a fully neural model (Figure~\ref{fig:method} (left)) for the downstream task. Such a model comprises the $f_e$ function and a continuous approximation $\tilde{g}$ of the symbolic component: 
$$\tilde{g}([z|w]) \approx g(f_m(z), f_m(w))$$
where $[\cdot | \cdot]$ represents the concatenation of two vectors.

Our method consists of two phases: in the first one, we train the neural model; in the second we copy and freeze the learned weights of $f_e$ (blue box in Figure~\ref{fig:method}) into the NeSy architecture. In this way, the NeSy method has to learn only $f_m$ (orange box) instead of the entire $f$. In other words, it has to learn only the mapping from embeddings to symbols. An exception is DSL~\cite{DSL}, which learns both $f$ and $g$ (green box). With the pretraining, it has to learn $f_m$ and $g$, allowing for much faster convergence.

Note that one has to use a neural architecture for pretraining which is task dependent. For instance, for the MNISTSum task, where the inputs are two single images, the concatenated embeddings are parsed by only a Feed Forward layer (Figure~\ref{fig:method} (left)). Instead, in the multi-digit variant of the task, the FF is substituted by a Recurrent Neural Network (RNN).

It is worth mentioning that the RNN has very competitive performance compared to other methods with similar accuracy, but much faster training. However, it generalizes poorly when increasing the number of digits with respect to the training samples. We experimentally observe that its performance does not degrade if the numbers have the same length as compared to the training set. In other words, the RNN struggles with out-of-distribution generalization. This can be seen in Table~\ref{tab:md}, where the performance of the RNN decrease significantly when passing from 2 digits(accuracy 92.9\%) to 4 digit (73.8\%) and 15 digit (18.9\%) numbers. This behaviour can be explained by errors produced at the symbolic level (i.e., at the level of the function $\tilde{g}$) due to the continuous approximation defined by the neural model. On the contrary, the mapping of the perceptions to the internal symbols ($f_e$) is generally correct since otherwise, the accuracy would be lower even with length 2 (the training numbers' length). Indeed, when analyzing the learned embeddings with t-SNE (Figure~\ref{fig:tsne}), we can clearly distinguish the various digits, confirming that the learned embeddings contain meaningful information to correctly classify the digits.

    \section{Experiments}
    In our experiments\footnote{The code and datasets used for the experiments are available in the supplementary materials.}, we focus on applying our approach to popular NeSy methods, namely: NeurASP~\cite{yang2020neurasp}, DeepProblog~\cite{manhaeve2018deepproblog}, DeepStochLog~\cite{winters2022deepstochlog}, Iterative Local Refinement ~\cite{ILR} and Deep Symbolic Learning~\cite{DSL}. We selected these methods due to their respective challenges: 

\begin{itemize}
    \item NeurASP (NAP), DeepProblog (DPL), and DeepStochLog (DStL) underperform when dealing with complex perception (e.g., CIFAR images);
    \item Iterative Local Refinement (ILR) often gets trapped in local minima;
    \item Deep Symbolic Learning (DSL) struggles to converge when dealing with problems with large amounts of latent symbols due to the large hypothesis space it needs to explore.
\end{itemize}

\label{chap:MNIST-SUM}

We test the aforementioned NeSy systems with and without the pretraining technique (PR) proposed in the paper, across a wide range of NeSy tasks. We observe that:

\begin{itemize}
    \item PR enhances convergence rates, leading to faster convergence across all the tested methods and tasks;
    \item PR eliminates local minima, leading to improved accuracy across most of the tested methods and tasks;
    \item PR allows the tested  NeSy methods to deal with complex perception inputs, where otherwise very poor results are obtained;
\end{itemize}

Moreover, due to the freezing of $f_e$ parameters, each epoch requires less time and memory for the backward pass of the back-propagation. Consequently, the overhead for the pretraining is negligible compared to the total gains obtained by adding it. For instance, the additional cost for the pretraining phase is 9 seconds (0.45 seconds per 20 epochs for the CNN training in Table~\ref{tab:table-sum}), while the time for training DSL has been reduced from 47.5 to 5.6 seconds in the MNISTSum task.

Overall, our experiments provide valuable insights into the limitations of the considered methods, showcasing how a pretraining phase overcomes specific challenges and improves their performance across diverse tasks.

\paragraph{Implementation Details.}
All the experiments were conducted with a machine equipped with an NVIDIA GTX 3070 with 12GB RAM. For digit classification, we use the same CNN as ~\cite{manhaeve2018deepproblog}, while for CIFARSum, we used a ResNet18~\cite{he2016deep} model, which provides a higher capacity than the CNN used for digit classification. This higher capacity is needed to correctly classify CIFAR images. The embedding $e$ obtained by $f_e(x)$ has size 84.  In the MNIST MultiDigitSum model pretrained for DSL$^\text{PR}$, the RNN has a hidden size of 80. Results are averaged over 10 runs. The t-SNE plots were obtained by applying the reduction technique directly on the embeddings $e$ and by reducing them to size 2. All the experiments have been obtained by adding our pretraining phase to the publicly available codebase of the tested methods. 

\paragraph{Evaluation Metrics.}
The standard metric used for the tackled task is the accuracy applied directly to the predictions of the final output (e.g., the sum value in the MNISTSum). This allows us to understand the general behaviour of the entire model. For the MNISTMultiDigitSum we also measured the fine-grained accuracy, that is, the average test accuracy of classifying the individual digits, e.g. if the output of the model is $12344$ and the ground truth is $12345$, then the fine-grained accuracy will be 0.8, since 4 digits out of 5 in the output are correct. Furthermore, we also measured the number of epochs required for convergence and the epoch time for every model we tested.

\begin{figure*}
    \centering
    \includegraphics[width=0.99\linewidth]{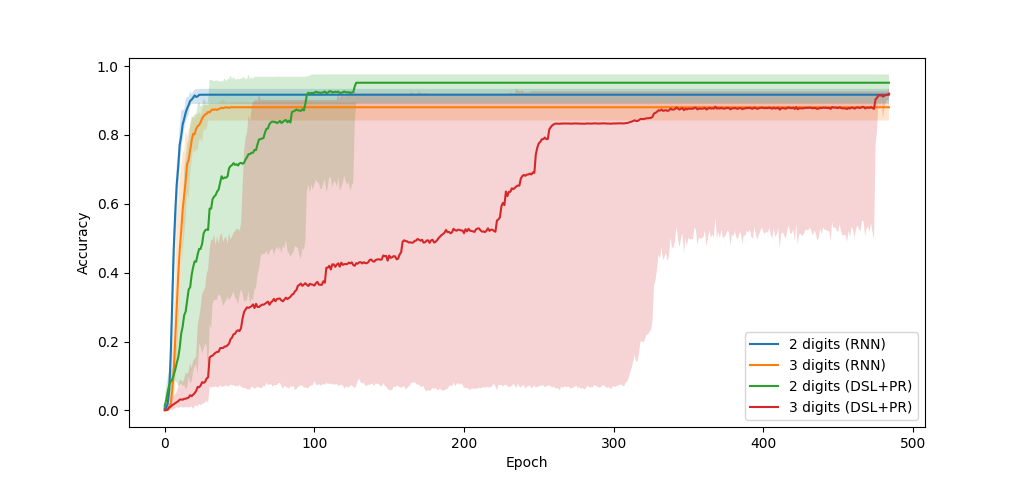}
    \caption{Minimum and maximum (transparent boundaries) and average accuracies (solid lines) obtained while training the RNN (on 2 or 3 digits) and DSL$^{\text{PR}}$. DSL convergence is slower but can generalize to longer sequences with higher results. The results are obtained across ten runs.}
    \label{fig:plot_multi}
\end{figure*}
\subsection{MNISTSum}
In the MNIST sum task, ~\cite{manhaeve2018deepproblog}, we are given triplets $(X, Y, Z)$, where $X$ and $Y$ are images of hand-written digits, while $Z$ is the result of the sum of the two digits, e.g., ($\digit{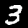}$, $\digit{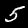}$, $8$). The goal is to learn to recognize MNIST images, using only the sum of the digits they represent as supervision. The results for this task are provided in Table~\ref{tab:table-sum}.
NeSy methods based on probabilistic semantics, such as DPL, NAP, and DStL, perform both accurately and fast on the MNISTSum task. Hence, DPL$^{\text{PR}}$, NAP$^{\text{PR}}$, and DStL$^{\text{PR}}$ show only marginal differences in performance from the non-PR versions, as not much space for improvement is left. Similarly, this phenomenon is also observed for DSL, which performs well with both PR and without it. However, in DSL, the number of symbols is not given apriori and an upper bound can be provided. A larger upper bound on the number of symbols leads to a lesser bias but a larger symbolic search space. We ran DSL on the MNISTSum task with an upper bound of 20 symbols, managing to get good results, even without PR. However, when the upper bound is moved to 1000, DSL does not converge in 1000 epochs (see DSL$_{\text{NB(1K)}}$). But when DSL is trained with the proposed pretraining method, i.e., DSL$_{\text{NB(1K)}}^{\text{PR}}$, it manages to show fast convergence with high accuracy. 
Finally, in ILR$^{\text{PR}}$ we observed higher averaged accuracy and much lower standard deviation. This is due to the ability of our pretraining strategy to deal with reasoning shortcuts~\cite{marconato2023neuro}, avoiding local minima.

\begin{table}[t]
\centering 
\begin{tabular}[t]{l|l|l|l|l}
       & Acc.(\%) & $\Delta_{Acc}$ & TE/\#E   & $\Delta_{\text{\#E}}$     \\ \hline
CNN                               & $98.2$\tiny{$\pm$0.8}  & - & $0.45s$/$20$ &  \multicolumn{1}{c}{-}   \\ \hline
NAP                               & $97.3$\tiny{$\pm$0.3}  & - & $109s$/$1$ &  \multicolumn{1}{c}{-}   \\
NAP$^{\text{PR}}$                 & $98.1$\tiny{$\pm$0.5}  & $0.8$ & $74s$/$1$ &  \multicolumn{1}{c}{-}   \\ \hline
DPL                               & $97.2$\tiny{$\pm$0.5}  & - & $367s$/$1$  & \multicolumn{1}{c}{-}   \\
DPL$^{\text{PR}}$                 & $97.6$\tiny{$\pm$0.3}  & $0.4$ & $284s$/$1$  & \multicolumn{1}{c}{-}   \\ \hline
DStL                              & $97.9$\tiny{$\pm$0.1}  & - & $8.2s$/$2$ &   \\ 
DStL$^{\text{PR}}$                & $97.4$\tiny{$\pm$0.3}  & -0.5 & $4.2s$/$1$ & -1  \\ \hline
ILR                               & $74.0$\tiny{$\pm$24.6} & - & $1.5s$/$10$ & \\ 
ILR$^{\text{PR}}$                 & $96.2$\tiny{$\pm$0.4}  & $22.2$ & $1.7s$/$1$ & -9 \\ \hline
DSL                               & $98.8$\tiny{$\pm$0.3}  & - & $0.95s$/$50$ & \\
DSL$^{\text{PR}}$                 & $98.6$\tiny{$\pm$0.1}  & -0.2 & $0.56s$/$10$ & -40\\  \hline
DSL$_{\text{NB(20)}}$             & $97.9$\tiny{$\pm$0.3}  & - & $0.99s$/$200$ & \\ 
DSL$_{\text{NB(1K)}}$             & $8.7$\tiny{$\pm$0.2}   & - & $7.4s$/$1000$ & \\ 
DSL$^{\text{PR}}_{\text{NB(1K)}}$ & $98.2$\tiny{$\pm$1.2}  & $89.5$ & $5.2s$/$30$ & -970 \\ \hline
\end{tabular} 
\caption{Results on MNISTSum task. $\Delta_{Acc}$ is the accuracy difference without and with PR. TE is the time required for each epoch, and \#E is the number of epochs.}
\label{tab:table-sum}

\end{table}

\begin{table}[t]
\centering 
\begin{tabular}[t]{l|l|l|l|l}
         & Acc (\%)          & $\Delta_{Acc}$ & TE/\#E  & $\Delta_{\text{\#E}}$ \\ \hline
    ResNet18          & $87.1$\tiny{$\pm$1.4}            &      \multicolumn{1}{c|}{-}          &  2.2s/100 & \multicolumn{1}{c}{-}\\  
    Embed2Sym          & $84.4$            &      \multicolumn{1}{c|}{-}          &  5908s/- & \multicolumn{1}{c}{-}\\  \hline
    DPL                & $31.4$\tiny{$\pm$1.1}      &               & $699s/20$ & \\  
    DPL$^{\text{PR}}$  & $75.7$\tiny{$\pm$1.9}      & $+44.3$       & $473.1s/1$ & -19 \\ \hline
    DStL               & $63.7$\tiny{$\pm$0.4}      &               & $10.5s/50$ & \\ 
    DStL$^{\text{PR}}$ & $90.6$\tiny{$\pm$1.2}      & $+26.9$       & $7.4s/10$ & -40\\ \hline
    DSL                & $8.4$\tiny{$\pm$2.7}      &               & $13.3s/300$ &\\ 
    DSL$^{\text{PR}}$  & $81.6$\tiny{$\pm$0.9}      & $+73.2$       & $2.7s/30$ & -270\\ \hline
    ILR                & $32.6$\tiny{$\pm$16.2}     &               & $134.0s/30$ & \\  
    ILR$^{\text{PR}}$  & $69.3$\tiny{$\pm$0.1}      & $+36.7$       & $133.0s/2$ & -28\\ \hline
    NAP                & $7.2$\tiny{$\pm$1.4}       &               & $1108s/1$ &\\ 
    NAP$^{\text{PR}}$  & $82.0$\tiny{$\pm$1.7}      & $+74.8$       & $283.0s/1$ & \multicolumn{1}{c}{-}\\ \hline
\end{tabular}
\caption{Results on CIFARSum task. $\Delta_{Acc}$ is the accuracy difference without and with PR. TE is the time required for each epoch, and \#E is the number of epochs.}
\label{tab:table-cifar}
\end{table}

\begin{table}[t]
\centering 
\begin{tabular}[t]{l|l|l|l|l}
       & Acc. (\%) & $\Delta_{Acc}$ & TE/\#E    & $\Delta_{\text{\#E}}$    \\ \hline
CNN & $96.6$\tiny{$\pm$0.8}  & & $0.56s$/$50$  &\\ \hline
DSL & $92.2$\tiny{$\pm$0.4}  & &$0.84s$/$9000$  &\\
DSL$^{\text{PR}}$  & $93.5$ \tiny{$\pm$0.2}  & $+1.3$ & $0.61s$/$200$ & -8800 \\
\hline
\end{tabular} 
\caption{Results on MNIST MultiOp task. $\Delta_{Acc}$ is the accuracy difference without and with PR. TE is the time required for each epoch, and \#E is the number of epochs.}
\label{tab:table-multiop}
\end{table}

\subsection{CIFARSum}
The CIFARSum task~\cite{EMB2SYM} is similar to MNISTSum, but the MNIST digits are replaced by CIFAR10 images, and each class is labelled with a number between 0 and 9. This task allows testing NeSy systems with a more complex perception task while keeping the reasoning complexity the same as the  MNISTSum task. The additional complexity in the perception space requires models with higher capacity.
To this end, we used a ResNet18 backbone for all the tested NeSy systems. Using a larger model increases the time required for every epoch but provides better perception classification performance. The results obtained on the CIFARSum task are summarized in Table~\ref{tab:table-cifar}. Just like the MNISTSum task, the accuracy is measured over the accuracy of the sum, obtained by adding the digit labels of the CIFAR10 images. 

All the tested methods suffered a significant loss of accuracy (in comparison to MNISTSum) when dealing with this task. However, each NeSy method shows significant improvements with the proposed PR method. Furthermore, all the NeSy systems we tested show significant improvements in the number of epochs required  for convergence and also in the run time of each epoch. Hence, getting a significant gain in total runtime. Notice that in Table~\ref{tab:table-cifar}, the row with ResNet18, shows the time required for pretraining. Again it can be checked that the overhead for getting the embeddings is marginal in comparison to the total improvement in the run time. Hence, this the CIFARSum task demonstrates that the PR enables both higher accuracy and faster convergence, in presence of complex perception taks.

\subsection{MNISTMultiOp}
In this section, we focus on the effect of PR when both perception and symbolic rules need to be learned. To date, very few such NeSy systems exist~\cite{OOD_Generalization}, and in this paper, we choose DSL for this analysis. Already in the MNISTSum task, DSL$^{\text{PR}}$ converges faster than DSL. We now investigate the effect of  PR on DSL, in the much more complex MNISTMultiOp task, as introduced in ~\cite{DSL}. This task extends the MNIST Sum task by incorporating additional operators as perception inputs. The perceptions for this task consist of two images from the MNIST dataset and a third symbol representing an operation selected from the set $ \{ +,-,\times, \div \}$. The images representing the operations are obtained from the EMNIST dataset~\cite{cohen2017emnist}, where we extracted the letter images for $A$, $B$, $C$, and $D$, corresponding to $+$, $-$, $\times$, and $\div$ respectively. It is important to note that the hypothesis space for this task encompasses a staggering $82^{400}$ possible symbolic functions.
The results obtained with DSL and DSL$^{\text{PR}}$ are shown in Table~\ref{tab:table-multiop}. Note that DSL required 9000 epochs to solve the task, an order of magnitude more than the 200 of DSL$^{\text{PR}}$. This is due to the huge exploration phase required to learn the symbolic function, especially in the first epochs when the underlying perceptions are mapped to the wrong symbols. DSL$^{\text{PR}}$  also gives better accuracy. The key reason for the improvements is that with the embeddings learned through PR,  DSL$^{\text{PR}}$   immediately learns to map the images to symbols. This saves significant exploration in the symbolic hypothesis space, which is otherwise done with wrong underlying mappings from the NN.

\begin{table}[t]
\centering
\begin{tabular}{l|l|l|l|l}
\multicolumn{5}{c}{Accuracy (\%)}                                                                                       \\

     & \multicolumn{1}{c|}{2}              & \multicolumn{1}{c|}{4}               & \multicolumn{1}{c|}{15}                                                       & \multicolumn{1}{c}{1000}          \\
\hline
RNN  & 92.9\tiny{$\pm$0.2} &                  73.8\tiny{$\pm$0.4} &       18.9\tiny{$\pm$1.6}&                                             0.0\tiny{$\pm$0.0}       \\
Embed2Sym  & 97.7 &                  91.6 &      66.4 &                       \multicolumn{1}{c}{-}       \\
NAP  & 93.9\tiny{$\pm$0.7} & \multicolumn{1}{c|}{T/O}             &\multicolumn{1}{c|}{T/O}                                                      & \multicolumn{1}{c}{T/O}           \\
DPL  & 95.2\tiny{$\pm$1.7} & \multicolumn{1}{c|}{T/O}             & \multicolumn{1}{c|}{T/O}                                                      & \multicolumn{1}{c}{T/O}           \\
DStL & 96.4\tiny{$\pm$0.1} & 92.7\tiny{$\pm$0.6}  & \multicolumn{1}{c|}{T/O}                                                      & \multicolumn{1}{c}{T/O}           \\ 
DSL$^{\text{CL}}$  & 95.0\tiny{$\pm$0.7} & 88.9\tiny{$\pm$0.5}  & 64.1\tiny{$\pm$1.5}                                           & 0.0\tiny{$\pm$0.0} \\
\hline
DSL$^{\text{PR}}$  & 95.3\tiny{$\pm$0.2} & 92.2\tiny{$\pm$0.2}  & 74.2\tiny{$\pm$1.1}                                           & 0.0\tiny{$\pm$0.0} \\
\hline
\multicolumn{5}{c}{Fine-grained Accuracy (\%)}\\
     & \multicolumn{1}{c|}{2}              & \multicolumn{1}{c|}{4}               & \multicolumn{1}{c|}{15}                                                       & \multicolumn{1}{c}{1000}          \\
\hline
RNN     & 97.5\tiny{$\pm$0.1} & 93.5\tiny{$\pm$0.3} & 89.5\tiny{$\pm$0.5} & 89.3\tiny{$\pm$0.5}\\
DSL$^{\text{CL}}$     & 97.9\tiny{$\pm$0.1} & 97.3\tiny{$\pm$0.1} & 96.7\tiny{$\pm$0.1} & 96.5\tiny{$\pm$0.1}\\ \hline
DSL$^{\text{PR}}$    & 98.2\tiny{$\pm$0.3} & 97.9\tiny{$\pm$0.2} & 98.1\tiny{$\pm$0.3} & 97.7\tiny{$\pm$0.4}\\
\hline
\end{tabular}
\caption{Results obtained on the MNIST MultiDigitSum task (for 2, 4, 15, 1000 digits). Fine-grained accuracy is the average test digit accuracy. T/O stands for timeout ($>$14400s).}
\label{tab:md}
\end{table}

\begin{table}[t]
\centering 
\begin{tabular}[t]{l|l|l|l|l}
       & Acc. & L. Tr. & L. Test & TE/\#E   \\ \hline
RNN & $50.9$\tiny{$\pm$0.3}  & 4 & 4 & $0.06s$/$40$ \\ \hline
DSL & $98.7$\tiny{$\pm$0.4}  & 4 & 20 &$0.33s$/$1k$ \\
DSL$^{\text{PR}}$  & $98.6$ \tiny{$\pm$0.2}  & 10 & 20 & $0.29s$/$200$ \\
\hline
\end{tabular} 
\caption{Results on Visual Parity task. L. Tr. and Test refers to the length of the input sequence in training and test set. $\Delta_{Acc}$ is the accuracy difference with and without PR. TE is the time required for each epoch, and \#E is the number of epochs.}
\label{tab:table-vp}
\end{table}
\subsection{Visual Parity}
In the Visual Parity task~\cite{shalev2017failures}, a list of images representing two classes is provided (MNIST images of zeros and ones). The task is to predict whether a specific class is repeated an odd number of times in the input sequence. Again, we focus on learning both rules and perceptions. Results are visible in Tab. \ref{tab:table-vp}. Without the pretraining, DSL converges with up to 4 images lists. When trained on longer sequences, it obtains an accuracy close to 0.5, i.e., equivalent to a random classifier. When PR is applied, DSL can scale up to 10 symbols at learning time. Finally, note that the RNN reaches a low accuracy. This is an important aspect of this experiment, since it proves that PR may aid learning, even when the pretrained model performs poorly.

\subsection{MNISTMultiDigitSum}
The last task we considered is the MNISTMultiDigitSum. It is a generalization of the MNISTSum, where the goal is to learn to perform the sum of numbers composed of multiple digits. Table~\ref{tab:md} shows the results achieved by the SOTA methods
This task is particularly difficult: probabilistic methods like DPL, NAP, and DstL reach good accuracies, but at inference, they can scale to at most to 4 digits. The RNN learns very efficiently (see Figure~\ref{fig:plot_multi}) and with very high performances, however, it does not generalize to longer sequences. Finally, DSL cannot converge unless trained with Curriculum Learning (DSL$^{\text{CL}}$), i.e., by first learning from examples with one digit numbers, and then learning from more complex numbers of length two. However, when trained with PR, i.e.,  DSL$^{\text{PR}}$ does not require Curriculum Learning. 

Notably, DSL$^{\text{PR}}$, as also DSL$^{\text{CL}}$, have an almost perfect Fine-grained accuracy, while the RNN performance degrades when the input digits arity grows.
DSL$^{\text{PR}}$ is also capable of learning how to solve the task even with a training set composed of 3 digits long numbers, although with a slower convergence compared to the training with 2 digits (see Figure~\ref{fig:plot_multi}).

\paragraph{Limitations and Future Works.}

Pretraining on the downstream task greatly helps NeSy systems in faster convergence and mitigation of local minima. 
However, 
this approach presents some limitations. For example, looking at Figure~\ref{fig:plot_multi} shows the slower convergence of DSL when training on three digits sum w.r.t. to the one on two digits. This suggests that when the task becomes harder both in perception and reasoning, the PR alone cannot push to fast convergence. It would be interesting to explore other similar techniques in the future. An example can be to use the model used during the pretraining phase not as a pretraining but as an auxiliary loss to be optimized during the training of the NeSy systems. This should regularize the training pipeline and can be further combined with Self-Supervised tasks such as   Rotation~\cite{feng2019self} or Jigsaw Puzzle~\cite{noroozi2016unsupervised} to learn complementary signals.

    
    
    \section{Conclusions}
    The paper introduces a general methodology that allows efficient training and scaling of a vast array of NeSy systems. The key insight of our methodology is that effective embeddings for perception inputs can be obtained with only weak supervision coming from the downstream task, without using any labelled data for the perception inputs or any symbolic reasoning. We investigate the proposed methodology on various state-of-the-art NeSy systems and on multiple NeSy tasks. Our experimental analysis shows that the embeddings generated through the proposed method form a much more informative input to the NeSy system than the raw input itself. We observed consistent and significant improvement in both accuracy and run-time on all the tested NeSy methods. Our methodology also enabled tasks with complex perception components (e.g., CIFAR10 images), which were previously unattainable for the tested NeSy systems. Furthermore, the additional computational overhead, required to learn the embeddings is marginal in comparison to the run-time gains obtained by using the embeddings.

    \paragraph{Acknowledgments}
    TC and LS were supported by the PNRR project Future AI Research (FAIR - PE00000013), under the NRRP MUR program funded by the NextGenerationEU.
    
    
\newpage
\bibliographystyle{named}
\bibliography{ijcai24}

\end{document}